\documentclass[10pt,twocolumn,letterpaper]{article}

\usepackage[pagenumbers]{cvpr}              %
\makeatletter
\@namedef{ver@everyshi.sty}{}
\makeatother
\usepackage{graphicx}
\usepackage{amsmath}
\usepackage{amssymb}
\usepackage{booktabs}
\usepackage{svg}

\usepackage[pagebackref,breaklinks,colorlinks]{hyperref}

\usepackage[capitalize,noabbrev]{cleveref}

\def\cvprPaperID{21} %
\def\confName{CVPR}
\def\confYear{2023}

\usepackage{siunitx} 
\usepackage{multirow}

\begin{document}

\title{SportsPose - A Dynamic 3D sports pose dataset}

\newcommand*\samethanks[1][\value{footnote}]{\footnotemark[#1]}
\author{Christian Keilstrup Ingwersen\thanks{Equal contribution}\hspace{4pt}$^{1,2}$ \and Christian Møller Mikkelstrup\samethanks \hspace{4pt}$^{1,2}$ \and Janus Nørtoft Jensen$^1$ \and Morten Rieger Hannemose$^1$ \and
Anders Bjorholm Dahl$^1$ \and
$^1$ Visual Computing, Technical University of Denmark \and $^2$ TrackMan A/S, Denmark \\
{\tt\small cin@trackman.com, s194345@student.dtu.dk, \{jnje, mohan, abda\}@dtu.dk}}

\maketitle

\begin{abstract}
Accurate 3D human pose estimation is essential for sports analytics, coaching, and injury prevention. However, existing datasets for monocular pose estimation do not adequately capture the challenging and dynamic nature of sports movements. In response, we introduce SportsPose, a large-scale 3D human pose dataset consisting of highly dynamic sports movements. With more than 176,000 3D poses from 24 different subjects performing 5 different sports activities, SportsPose provides a diverse and comprehensive set of 3D poses that reflect the complex and dynamic nature of sports movements. Contrary to other markerless datasets we have quantitatively evaluated the precision of SportsPose by comparing our poses with a commercial marker-based system and achieve a mean error of 34.5 mm across all evaluation sequences. This is comparable to the error reported on the commonly used 3DPW dataset. We further introduce a new metric, local movement, which describes the movement of the wrist and ankle joints in relation to the body. With this, we show that SportsPose contains more movement than the Human3.6M and 3DPW datasets in these extremum joints, indicating that our movements are more dynamic. %
The dataset with accompanying code can be downloaded from our website \footnote{\url{http://christianingwersen.github.io/SportsPose}}. We hope that SportsPose will allow researchers and practitioners to develop and evaluate more effective models for the analysis of sports performance and injury prevention. With its realistic and diverse dataset, SportsPose provides a valuable resource for advancing the state-of-the-art in pose estimation in sports.
 
\end{abstract}

\section{Introduction}
\label{sec:intro}
Monocular 3D human pose estimation is a blooming topic enabling human-computer interaction with applications in biomechanics~\cite{nishimura2020a}, entertainment~\cite{zhang2012a}, sports~\cite{hwang2017a,zecha2018a, scott2017a, rematas2018a}, and many more. Recent methods have shown impressive performance with in-the-wild methods achieving mean per joint precision errors (MPJPE) of less than 8~cm \cite{dynaboa, kocabas2021pare, meshgraphformer, romp}. 

Large datasets enable advancing the state-of-the-art for pose models, however acquiring 3D human pose datasets is a cumbersome and expensive process that usually requires a commercial motion capture system based on inertial measurement units (IMU) or optical markers  ~\cite{HumanEva, h36m, 3dhp, totalcapture}. This complexity tends to constrain the capture of human pose datasets to controlled lab environments with a minimal number of different subjects. Having markers attached to the body can also be impractical, affecting the subject's ability to move freely and potentially reducing the generalization of models trained on the data, as the models can start to rely on the visible markers to estimate the pose. 

Because of these issues markers are not desirable in a dataset for vision-related learning problems, and a markerless capture system is preferred instead. Various 3D human pose datasets, recorded in outdoor environments \cite{aspset} and controlled indoor lab setups~\cite{panoptic}, are available. However, existing markerless datasets lack a quantitative analysis to validate their accuracy, which raises concerns regarding the quality of the data considered ground truth.

The 3DPW dataset \cite{3dpw} addresses the issue of visible markers by utilizing an IMU-based system, which allows most sensors to be concealed under clothing. The IMU data is then aligned with video data from a mobile camera. To evaluate the effectiveness of this method, a quantitative analysis is performed using the TotalCapture dataset \cite{totalcapture}, which contains both optical marker and IMU data. However, since the TotalCapture data set is recorded in a different environment than the rest of the 3DPW dataset, it is unclear whether the measured error accurately reflects the expected error. Despite this limitation, the reported mean per joint precision error on the TotalCapture dataset is 26~mm. In contrast, we introduce SportsPose, a markerless human 3D pose dataset, which includes a quantitative analysis of the estimated markerless poses. To validate the accuracy of our dataset, we compare it with a commercial marker-based motion capture system in the same domain. Our results indicate a precision on par with the 3DPW dataset but measured in the same domain as the data was captured.

\begin{figure*}[t]
\centering
\includegraphics[width=0.9\textwidth]{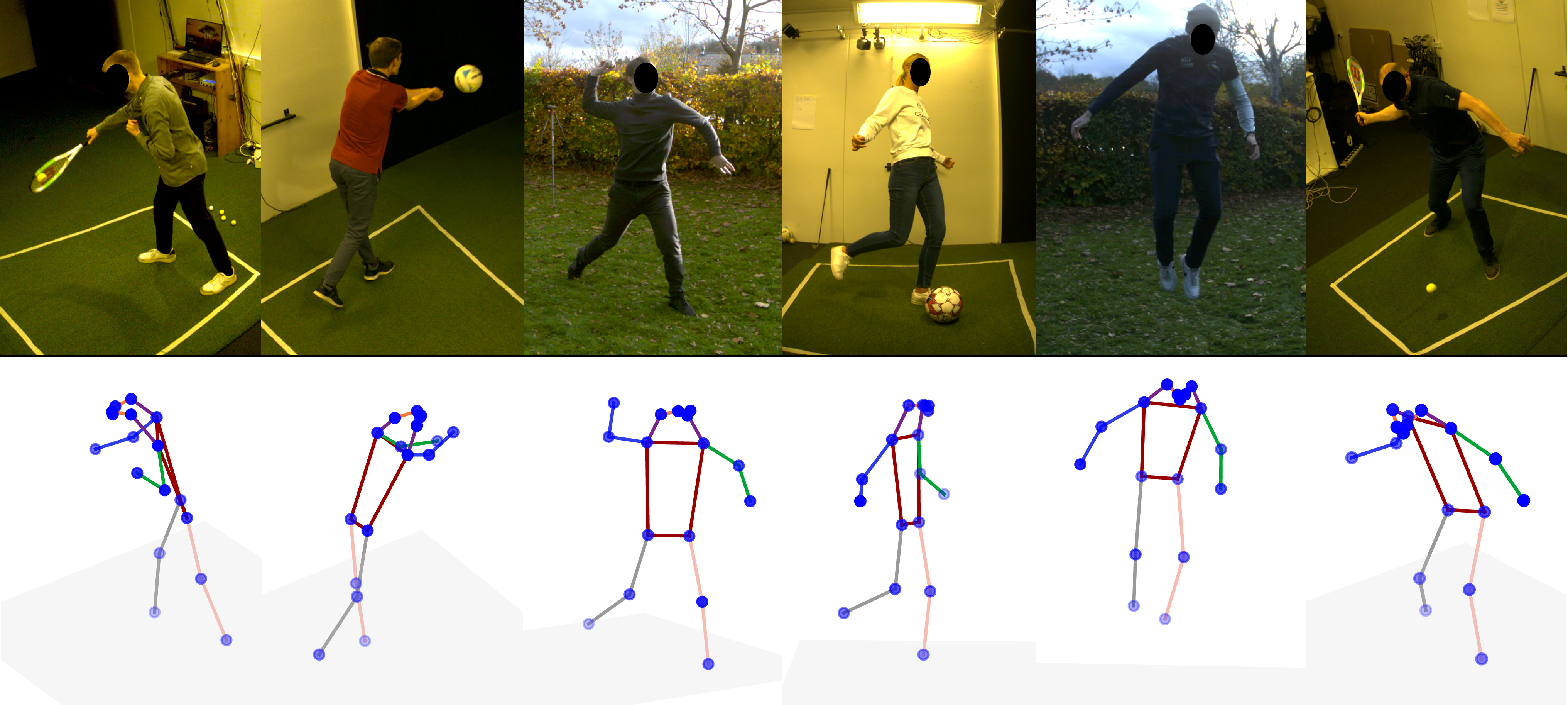}
\caption{Examples from each of the activities in the dataset with corresponding 3D poses. The samples are from both indoor and outdoor captures. It should be noted that the subjects in the figure are anonymized which they are not in the released data.}
\label{fig:data_samples}
\end{figure*}

With SportsPose, we present a markerless 3D human pose dataset with data from a total of 24 subjects in indoor and outdoor environments. 
We include five sports activities namely, soccer, volleyball, jump, baseball pitch, and tennis. These activities have been chosen because they are highly dynamic movements, including a large range of motion while being possible to perform in a constrained capturing volume. Samples from the dataset of different activities and different subjects can be seen in \cref{fig:data_samples}. The subjects in \cref{fig:data_samples} are anonymized, but in the available licensed dataset, they are not.  

 A calibrated and hardware-synchronized setup of 7 color cameras recorded the sequences of poses at a rate of 90 Hz. Using a pre-trained 2D pose detector \cite{hrnet}, a 2D pose was predicted for each image, yielding multiple 2D poses from different views. We obtained a range of 3D point candidates by triangulating from multiple camera subsets. A graph-based approach improved temporal continuity, followed by Butterworth smoothing, which reduced the candidates to a smooth sequence of 3D poses for all frames. The accuracy of the estimated 3D human movements was evaluated on a separate set of videos by comparing them with a commercial marker-based motion capture system that recorded the same volume. This comparison revealed a mean error of 34.5 mm across the separate set of videos.

Current models fail to accurately predict joint locations for dynamic sports movements \cite{ingwersen2023evaluating}. There is no existing sports dataset with such dynamic movements, variability in poses, and rigorous accuracy evaluation as SportsPose. Our goal with SportsPose is to encourage research that advances monocular 3D models.

To summarize, our contribution is:
\begin{itemize}
    \item The SportsPose dataset -- a large markerless human 3D pose dataset.
    \item Quantitative analysis of the accuracy of the reference poses.
    \item Dynamic sports movements of 24 subjects.
    \item An easily scalable motion capture system for future dataset extensions.
\end{itemize}

\section{Related Work}
\label{sec:related_word}
\subsection{3D human pose datasets}
There have been numerous efforts to build large 3D human pose datasets to train and evaluate monocular 3D human pose estimation models. Notable examples of such datasets include HumanEva \cite{HumanEva}, TotalCapture \cite{totalcapture}, Human3.6M \cite{h36m}, and CMU Panoptic Studio \cite{panoptic}. These datasets have been instrumental in advancing the state-of-the-art in monocular 3D human pose estimation. 
Acquisition of accurate 3D human pose data has previously been constrained to controlled lab setups with a small and fixed capturing volume \cite{h36m, panoptic, HumanEva, totalcapture}. Human3.6M \cite{h36m}, HumanEva \cite{HumanEva}, and TotalCapture \cite{totalcapture} all use an optical tracking system with infrared cameras and reflective markers mounted on all of the subjects. Acquiring motion capture data with these marker-based optical systems is considered to be the golden standard for accurate motion capture and are the systems used for research in biomechanics \cite{sandbakk2012influence, mirek2007assessment}. 

There are certain limitations to using a marker-based system, particularly for highly dynamic movements such as those in sports, as markers can cause discomfort and potentially impede the subjects performance. Additionally, the presence of optical markers can create an artificial environment that may not reflect real-world scenarios. A concern is that models may learn the appearance of these markers for estimating the pose, leading to poor generalization to markerless situations. An alternative to the optical marker-based systems is an IMU-based system, as used in the 3DPW dataset \cite{3dpw}. Such systems allows for a less constrained environment and the option to hide some of the sensors under the subjects' clothing, but may have issues with measurements drifting. The 3DPW dataset solved this issue by mounting IMU sensors on the subject and correlating IMU sensor data with video from a mobile camera to obtain accurate 3D poses of subjects in various environments, making it a truly ``in the wild" dataset. However, a downside of their approach is that the subject needs to wear visible IMU sensors, and the lack of available ground truth data makes it difficult to evaluate the algorithm's performance in aligning IMU and video data.

The CMU Panoptic dataset \cite{panoptic} was able to capture 3D pose data without relying on markers or IMU sensors. Instead, they utilized a multi-camera setup to detect 2D poses and triangulate the corresponding 3D pose. However, their setup is quite extensive, requiring 480 industrial-grade cameras and 10 Kinect 2 sensors, making it difficult to reproduce. In contrast, our SportsPose system employs a similar approach but with only seven cameras, which makes it more accessible and portable to new capture locations. To ensure the system remains accurate while being portable we have conducted a quantitative comparison with an optical marker-based system. %

Other methods for developing a flexible markerless capturing system have been proposed, including ASPset-510 \cite{aspset}, which employs three consumer-level cameras and manual time synchronization to construct an outdoor human sports pose dataset. However, no quantitative analysis of the dataset's accuracy is provided in ASPset-510. Our study revealed that more than three cameras were necessary for sports movements due to frequent self-occlusions. We also discovered that a hardware-based frame, not just time synchronization, was required for our movements, as joints could move excessively between frame exposures. Seven cameras with hardware synchronization proved to be a good compromise between system accuracy, cost, and flexibility when developing SportsPose.

Markerless motion capture systems are commercially available \cite{CapturyMarkerless, TheiaMarkerless, simiMarkerless} and have been employed to construct datasets for deep learning. The MPI-INF-3DHP~\cite{3dhp} dataset utilized a commercial markerless solution \cite{CapturyMarkerless} to capture diverse poses without markers. While it contains motion capture data for eight subjects in natural clothing, it was captured in a lab environment with a green screen. In contrast, SportsPose provides a dataset with a substantial number of subjects in natural settings, and its accuracy has been evaluated using a precise optical reference system. A summary of the motion capture datasets discussed in this section is presented in \Cref{tab:dataset_summary}.

\subsection{Monocular 3D human pose models}
The subject of monocular 3D human pose estimation has been widely explored, with two main approaches to inferring the 3D pose. One approach is a single-stage method, which employs parametric body models to predict both the shape and pose of a subject directly from an input image or video, such as those found in \cite{xu2020ghum, smplify, dynaboa, spin, hmr, metro}. The other approach is a two-stage method, which uses either a ground truth or a predicted 2D pose to estimate the corresponding 3D pose of a subject, as seen in \cite{pavllo:videopose3d:2019, Zhang_2022_CVPR, shan2022p, chen2020anatomy}. Each approach has its benefits and drawbacks, but if only the pose is relevant, the two-stage methods are considered the most accurate~\cite{pavllo:videopose3d:2019}. Additionally, two-stage methods allow for more temporal information to be included as the lifting module only takes 2D poses as input rather than full image frames. With SportsPose, our focus has been on advancing accurate sports pose estimation rather than shape estimation. We have released camera calibrations to allow for ground truth 2D poses to be used in two-stage approaches. If shape information is needed, it can potentially be obtained using a motion capture body solver like MoSh~\cite{Loper:SIGASIA:2014}.

\section{Motion capture system}
The system we built to capture the SportsPose dataset consists of seven hardware-synchronized industrial cameras capturing at 90Hz with a resolution of $1920 \times 1200$.  The cameras are mounted around a capturing space of two by two meters with some cameras in the ceiling and others mounted at chest height. The system is calibrated using a board with six ArUco patterns \cite{aruco}, first obtaining a linear estimate using Zhang's method \cite{zhang2000a} followed by non-linear bundle adjustment, resulting in a mean re-projection error of 0.8 pixels. 

\subsection{Triangulation procedure}
\begin{figure}[h]
    \centering
    \includegraphics[width=\linewidth]{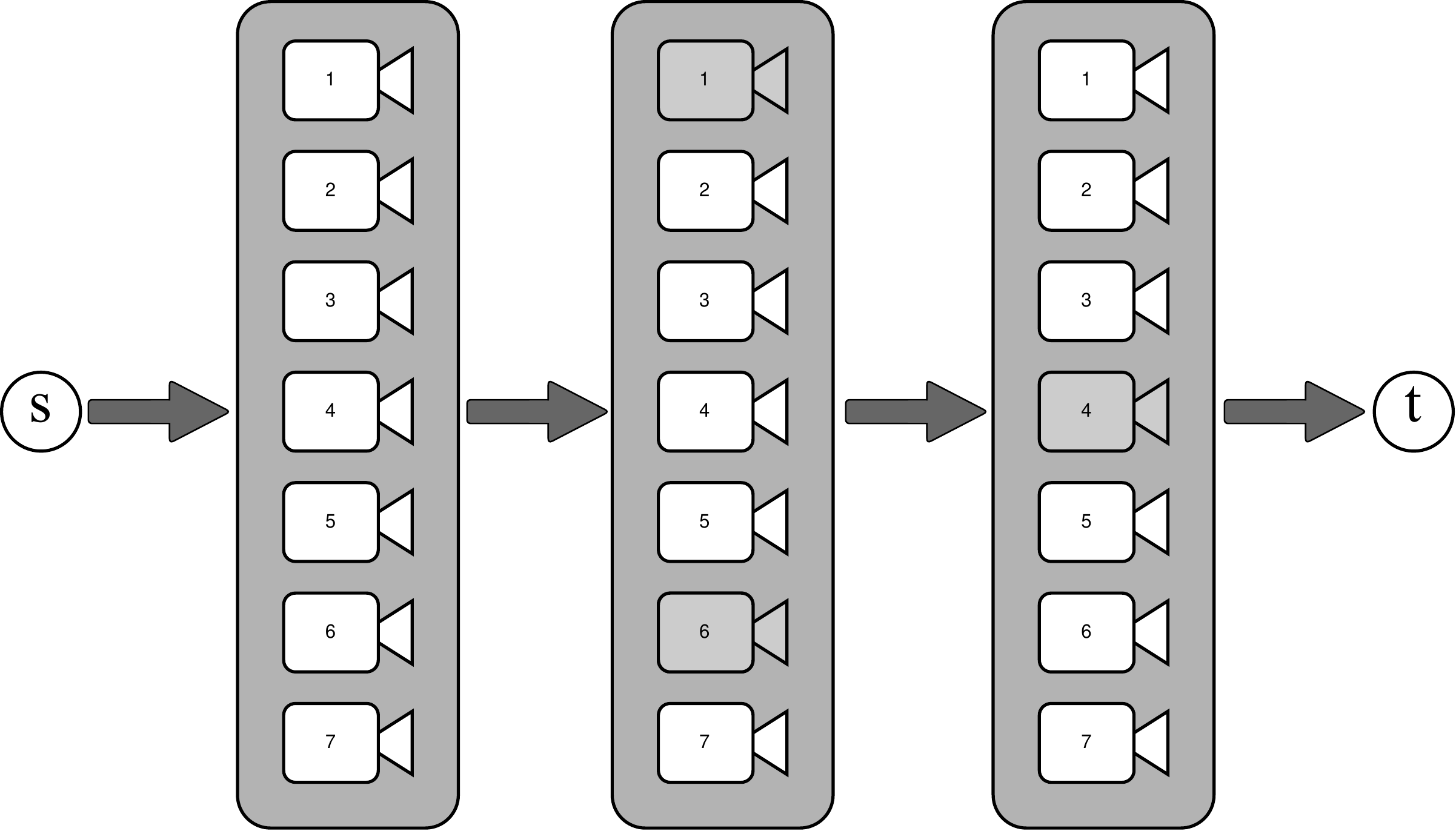}
    \caption{All possible subsets with a minimum of two cameras are connected densely, with each layer corresponding to one frame. Up to two cameras, and thus their subsets, can be removed if the pose estimator has low confidence, here shown as greyed-out cameras.}
    \label{fig:camera_dag}
\end{figure}

To estimate the 3D human pose, we utilize the 2D pose detector HRNet \cite{hrnet} to predict an initial set of 2D joints from all the camera views. The specific HRNet model used is trained on the COCO 2D pose dataset \cite{cocodataset}, and SportsPose's markerset is thus identical to the one in COCO. With the predicted 2D joint locations we triangulate a linear estimate of the 3D joint positions, which we refine using non-linear optimization. This estimate of the 3D joint location can potentially be erroneous due to noisy predictions from the 2D estimator that may have jitter, joint swaps, and other errors~\cite{ronchi2017a}. To ensure temporally coherent 3D joint positions and correct for potential erroneous predictions, we use information from previous and future frames to refine our estimate of the current joint locations. This is, inspired by ASPset-510~\cite{aspset}, done by constructing the set of possible estimates; all 3D points that can be triangulated using two or more cameras for a total of $\sum_{i=2}^K$$K \choose i$ point candidates~\cite{aspset}. Thereby, one can let each point candidate for each time stamp be a vertex in a directed acyclic graph as illustrated in \cref{fig:camera_dag}. We let each vertex be connected to all vertices in the following time step. Assuming little movement between frames, we try to minimize the distance moved between frames for each joint, and so the edge weights $w_{ij}$ between two vertices, $v_i$ and $v_j$, becomes,

  \begin{equation}
    w_{ij} = ||v_i-v_j||_2.
  \end{equation}

  Using dynamic programming, one can efficiently find the shortest path in the graph, giving the 3D locations for all time steps for each joint. To utilize the information given by the pose estimator, we use the 2D joint confidence to remove up to two cameras and the corresponding nodes that use this camera from the graph for every frame, as illustrated in \cref{fig:camera_dag}. We settled for up to two cameras since this struck a good balance between removing the most unconfident cameras and allowing the graph-approach to find a smooth sequence of poses.
  The points picked out by the graph-approach are additionally smoothed using a Butterworth filter~\cite{williams2006a, parks1987a}, which is widely used within biomechanics~\cite{winter2009a}. The filter is designed as a fourth order filter with a cutoff frequency of 6~Hz, since the majority of human movement is captured at this frequency~\cite{winter1974a}.

\section{SportsPose dataset}
\label{sec:data_overview}
With the described multi-camera setup we have collected the SportsPose dataset, consisting of a total of $191,948$ 3D poses from highly dynamic sports movements from 24 subjects, currently making this the 3D pose dataset with the most subjects, see \cref{tab:dataset_summary}. The 3D poses in the dataset are distributed with $149,580$ poses in an indoor environment, $27,000$ outdoors and $15,368$ poses in an indoor environment with optical markers on the subjects used for our quantitative quality assessment in \cref{sec:data_quality}.

\subsection{Dataset}
The ease of use of our system has allowed us to scale the number of subjects to a total of 24 with, 3 female and 21 male participants, all wearing natural clothing and no markers attached. We have further captured data in both an outdoor and indoor environment where two of the subjects appear in both the indoor and outdoor settings, which allows ablations of a model's performance in different environments. 
Each subject is recorded performing 5 repetitions of 5 short sports-related activities, resulting in a total of $191,948$ poses with corresponding images from 7 cameras, totaling 1.5 million frames. Of the $191,948$ poses, $15,368$ are used for the quality assesment and the subjects here have visible markers on the body. The activities in the dataset are baseball pitch, jump, tennis, volleyball, and soccer. They were chosen to allow the subjects to perform a wide variety of poses within the volume, allowing fast-moving joints in both the upper and lower part of the body. The subjects were informed of the movements and how to perform them but allowed creative freedom to perform them as they wanted. 

To summarize the contents of the SportsPose dataset and to compare it to other current 3D pose datasets, we provide an overview in \cref{tab:dataset_summary}. It can be seen that SportsPose is the largest motion capture dataset in terms of the number of subjects and the fourth largest dataset in terms of the number of 3D poses behind the Human3.6M, CMU Panoptic, and TotalCapture datasets \cite{h36m, panoptic, totalcapture}. Human3.6M \cite{h36m} is a marker-based dataset, while the CMU Panoptic dataset \cite{panoptic} is a markerless system similar to ours but with more cameras. In the CMU Panoptic dataset, they constructed an indoor dome with more than 400 low-resolution cameras and 31 high-resolution cameras to capture their dataset \cite{panoptic}. This can be considered the golden standard in markerless datasets but it also completely removes the flexibility of moving the system to more natural environments. This makes SportsPose the second largest publicly available markerless dataset, the dataset with the highest framerate data, and  the dataset with the most subjects.

\sisetup{table-align-text-post=false, table-space-text-post ={MM},table-number-alignment=center-decimal-marker}
\begin{table*}[t]
\centering
\newcommand{\worst}[1]{\textbf{#1}}
\newcommand{\mr}[1]{{\multirow{2}{*}{\text{#1}}}}
\begin{tabular}{l *{10}{S[table-format=3.1]}}

\toprule
& \text{Marker-} & \text{Quality} & \mr{Sync} &\mr{Subjects} & \mr{Poses} & \mr{Environment}& \mr{Views} & \mr{FPS} &\mr{Frames}  \\
& \text{less} & \text{evaluation} & & & & & &\\ 
\midrule
\text{Human3.6M} \cite{h36m} & $\times$ & \checkmark & \text{hw} & 11 & \text{900K} & \text{Indoor} & 4 & 50 & \text{3.6M} \\
\text{MPI-INF-3DHP} \cite{3dhp} & \checkmark & $\times$ & \text{hw} & 8 & \text{93K} & \text{Indoor} & 14 & \text{N/A} & \text{1.3M} \\
\text{3DPW} \cite{3dpw} & $\times$ & \checkmark & \text{sw} & 7 & \text{49K} & \text{In- \& outdoor} & 1 & 30 & \text{51K} \\ 
\text{HumanEva-I} \cite{HumanEva} & $\times$ &  \checkmark & \text{sw} & 6 & \text{78K} &  \text{Indoor} & 7 & 60 & \text{280K} \\
\text{HumanEva-II} \cite{HumanEva} & $\times$ & \checkmark & \text{hw} & 6 & \text{3K} & \text{Indoor} & 4 & 60 & \text{10K} \\
\text{TotalCapture} \cite{totalcapture} & $\times$ & \checkmark & \text{hw} & 5 & \text{179K} & \text{Indoor} & 8 & 60 & \text{1.9M} \\
\text{CMU Panoptic} \cite{panoptic} & \checkmark & $\times$ & \text{hw} & 8 & \text{1.5M} & \text{Indoor} & 31 & 30 &\text{46.5M} \\
\text{ASPset-510} \cite{aspset} & \checkmark & $\times$ & \text{sw} & 17 & \text{110K} & \text{Outdoor} & 3 & 50 & \text{330K} \\
\text{SportsPose (ours)} & \checkmark & \checkmark & \text{hw} & 24 & \text{177K}  & \text{In- \& outdoor} & 7 & 90 & \text{1.5M}  \\
\bottomrule
\end{tabular}
    \caption{Summary statistics of public pose datasets. Sync refers to whether the cameras are hardware (hw) or software (sw) synchronized. It can be seen that SportsPose is the second largest markerless dataset and the dataset with the highest framerate and largest amount of subjects.}
    \label{tab:dataset_summary}
\end{table*}

\subsection{SportsPose statistics}
For a thorough analysis of the poses and movements in the SportsPose dataset and to be able to compare it to existing datasets, we have calculated a series of statistics for SportsPose, 3DPW \cite{3dpw}, and Human3.6M \cite{h36m}. We compare to 3DPW and Human3.6M as they are the most commonly used datasets for developing new 3D human pose estimation methods and represent the current go-to dataset in respectively lab scenarios and in the wild scenarios.

\begin{figure}[t]
    \centering
    \includegraphics[width=0.99\linewidth]{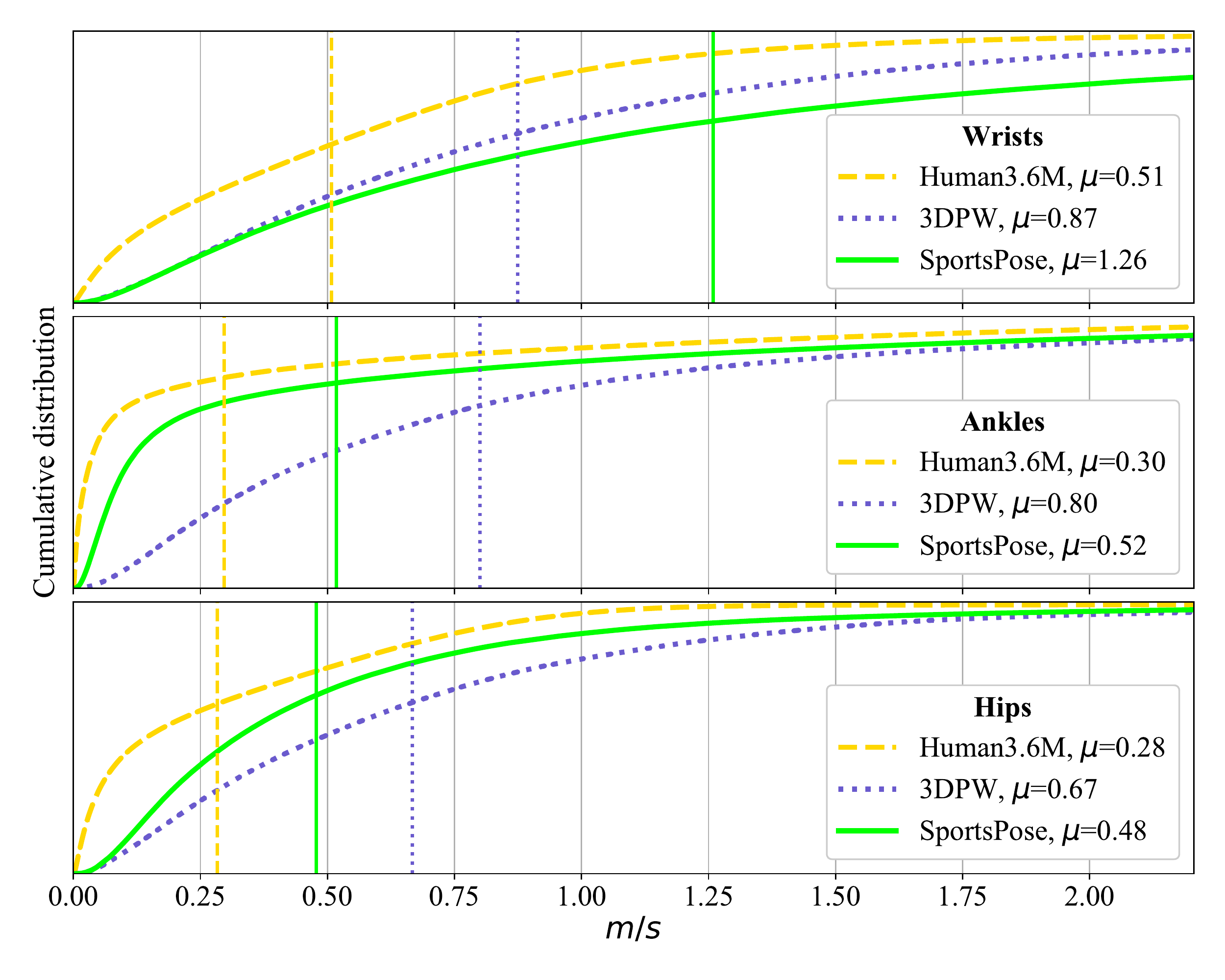}
    \caption{Comparison of the wrist, ankle, and hip speed as a cumulative distribution function for Human3.6M, 3DPW, and SportsPose. Lower lines indicate higher speeds. The mean speed for each of the datasets are indicated by a vertical line.}
    \label{fig:hist_speed}
  \end{figure}

\begin{figure}[t]
    \centering
    \includegraphics[width=0.99\linewidth]{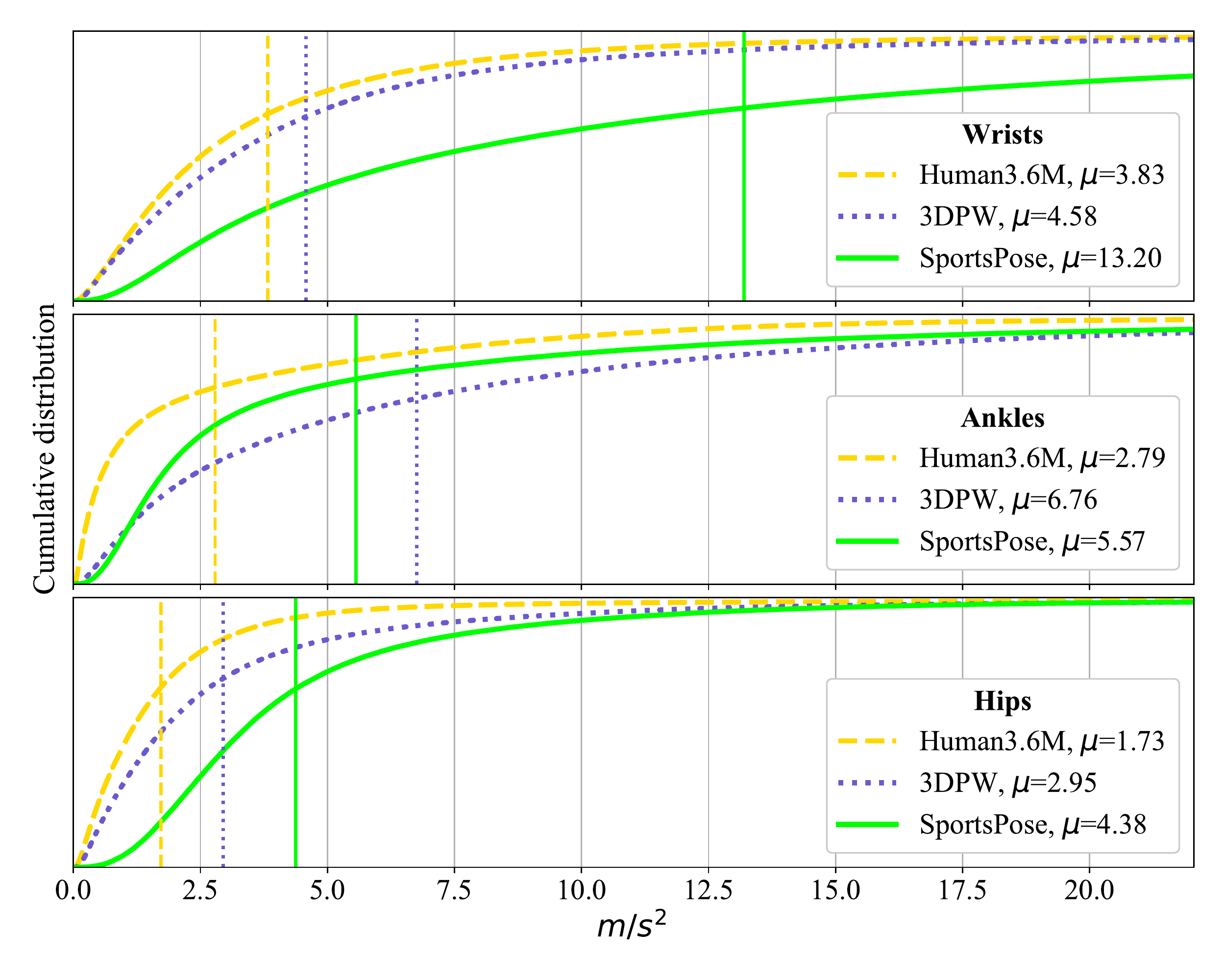}
    \caption{Comparison of the wrist, ankle, and hip acceleration as a cumulative distribution function for Human3.6M, 3DPW, and SportsPose. Lower lines indicate higher accelerations. The mean acceleration for each of the datasets are indicated by a vertical line.}
    \label{fig:hist_accel}
  \end{figure}

To investigate how dynamic the movements in the datasets are, we have computed the speed and acceleration for all wrists, ankles, and hips in the datasets. The cumulative distribution functions of these are in \cref{fig:hist_speed,fig:hist_accel}. From these distribution functions it becomes clear that the movements in the SportsPose dataset differ from the movements in 3DPW and Human3.6M in terms of speed and acceleration, which is to be expected since we specifically target dynamic sports movements. Inspecting the plots further, we see that wrists from SportsPose have the fastest speed and acceleration of the three datasets. This makes sense since many sports-related movements are short bursts of high acceleration resulting in fast movements, like throwing a ball. Additionally, we also see high speed in the ankles and hips, as opposed to Human3.6M. The 3DPW dataset has the fastest speeds in both ankles and hips, which most likely is due to their larger recording volume allowing the subject to move freely around as opposed to Human3.6M and SportsPose where the capturing volume is fixed in size and space.

\Cref{fig:hist_speed,fig:hist_accel} tell us that SportsPose contains fast movements but we cannot conclude anything related to the variety of poses based on this. With SportsPose we contribute with a dataset not only with fast movements but also a large variety of movements and poses. To demonstrate this, we want to measure how much volume the joints move through around the subject. We propose a new measure, \textit{local movement} to quantify joint movement. 

\begin{figure}
    \centering
    \includegraphics[width=\linewidth]{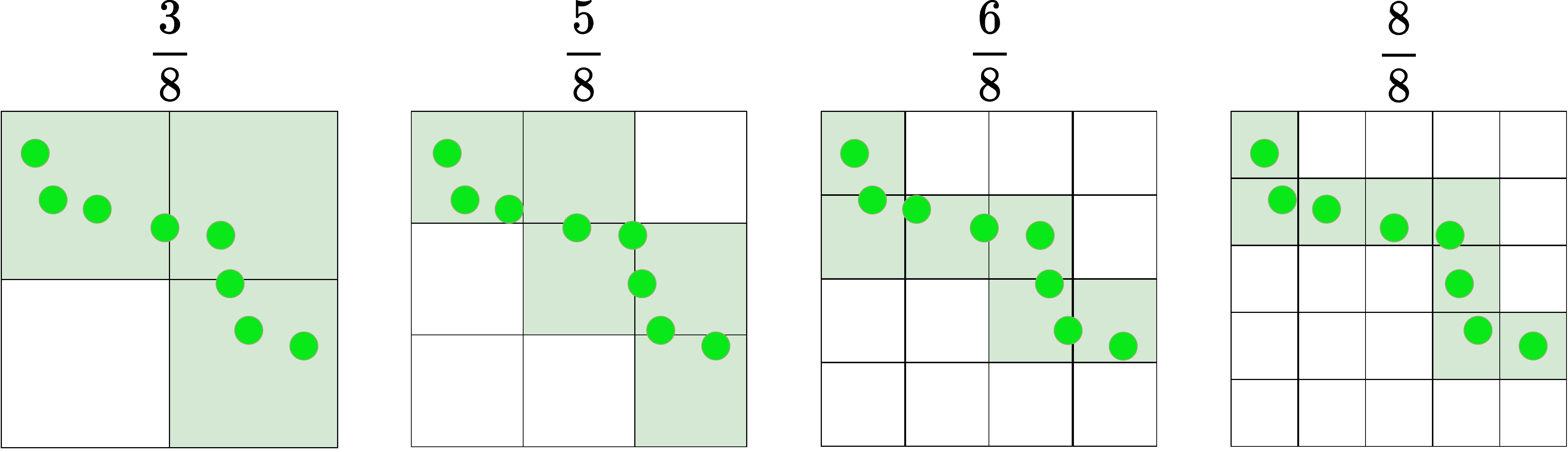}
    \caption{An example of local movement visualized in 2D. The joints are shown in dark green and the visited voxels in a light green, with the corresponding cover ratio on top. It can be seen that the same number of joints, occupy a larger number of voxels as the resolution of the voxel grid is increased.}
    \label{fig:local_movement}
\end{figure}

\begin{figure}[t]
  \centering
  \includegraphics[width=\linewidth]{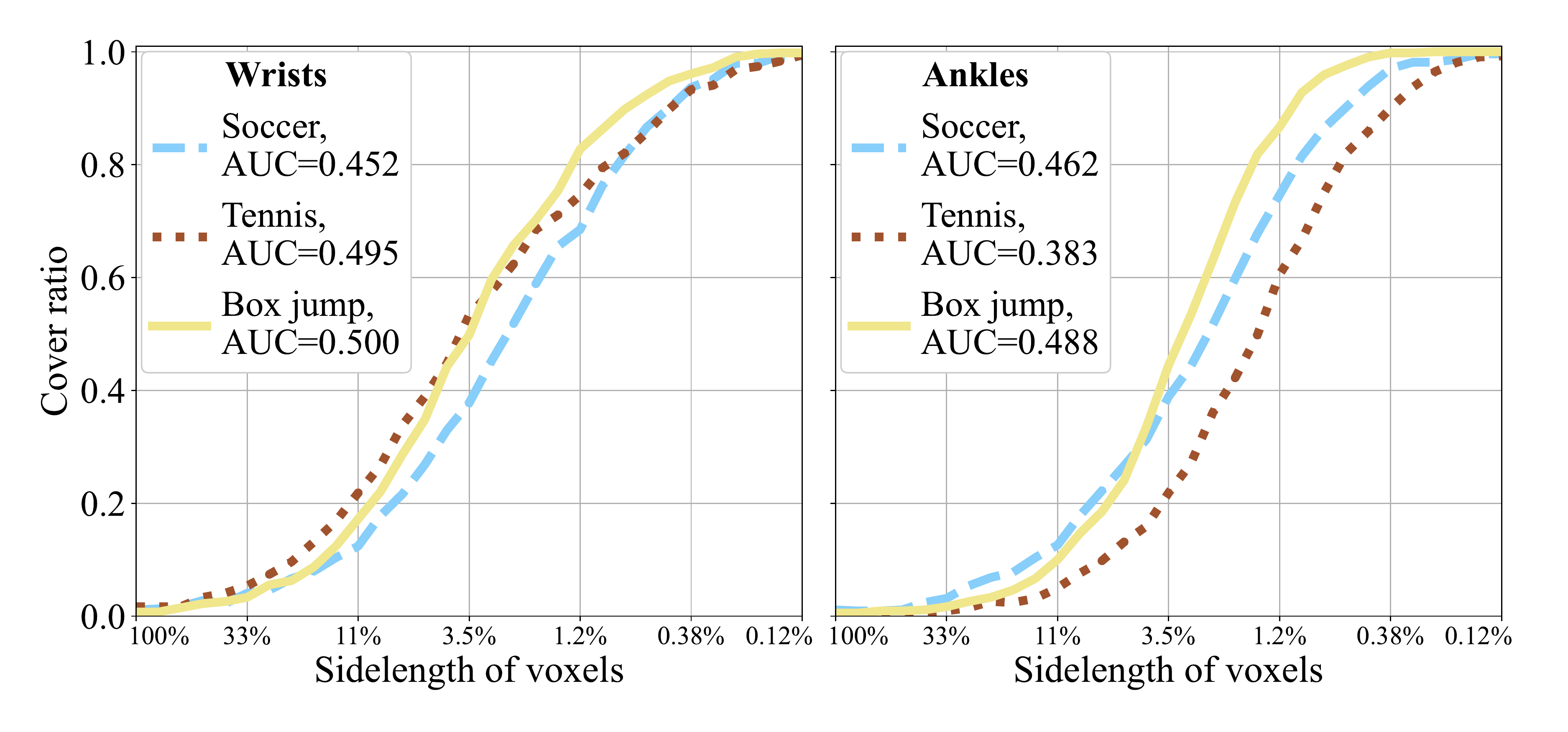}
  \caption{Fraction of unique voxels covered to the number of joints in a local coordinate system as a function of the size of the voxels. The local movement plots are a measure for movement of the wrists and ankles, here shown for a single sequence of three SportsPose activities.}
  \label{fig:LM_plot_example}
\end{figure}

\textit{Local movement} only considers the extremum of the body joints i.e.\ the wrists and ankles, which have the most freedom to move relative to the body. To capture the movement of these joints relative to the rest of the body, we do a frame-wise rotation and translation for a change of coordinate system. For the wrists, we have the new origin at the shoulder with the $x$-axis aligned with the shoulders and the hip-to-shoulder vector lying in the $xz$-plane, similarly for the ankles centered at the hips and its $x$-axis aligned with the hips. To exploit symmetry, the left-hand joints are mirrored and placed in the same coordinate system as the right-hand joints. To figure out how much volume the wrists and ankles move through, we place a grid of voxels in the new coordinate system with its sides aligned to the basis vectors. By finding the \emph{cover ratio}, i.e.\ the number of unique voxels occupied by wrists or ankles divided by the total number of frames, divided by two to account for mirroring, we get a quantity that indicates how much movement was performed locally to the subject. The larger the side length of the voxels is when the cover ratio approaches $1$, the more volume is covered throughout the movement. This is illustrated in a 2D projection in \cref{fig:local_movement}. By calculating the cover ratio for $n$ voxel side lengths log-spaced from $1$ to $1/1000$ of the length of an arm or a leg, and finding the area under the curve divided by $n$, we get a metric that can be used to compare two sets of poses. This is illustrated for three sequences from SportsPose in \cref{fig:LM_plot_example}. Here we find that tennis has more movement in the wrists than soccer, but less ankle movement, reflecting the movements required in the activities. We also see that the box jump has a larger area under curve than both of the others, which is to be expected as both the ankle and wrist joints move a lot when jumping. 

\begin{figure}[t]
  \centering
  \includegraphics[width=\linewidth]{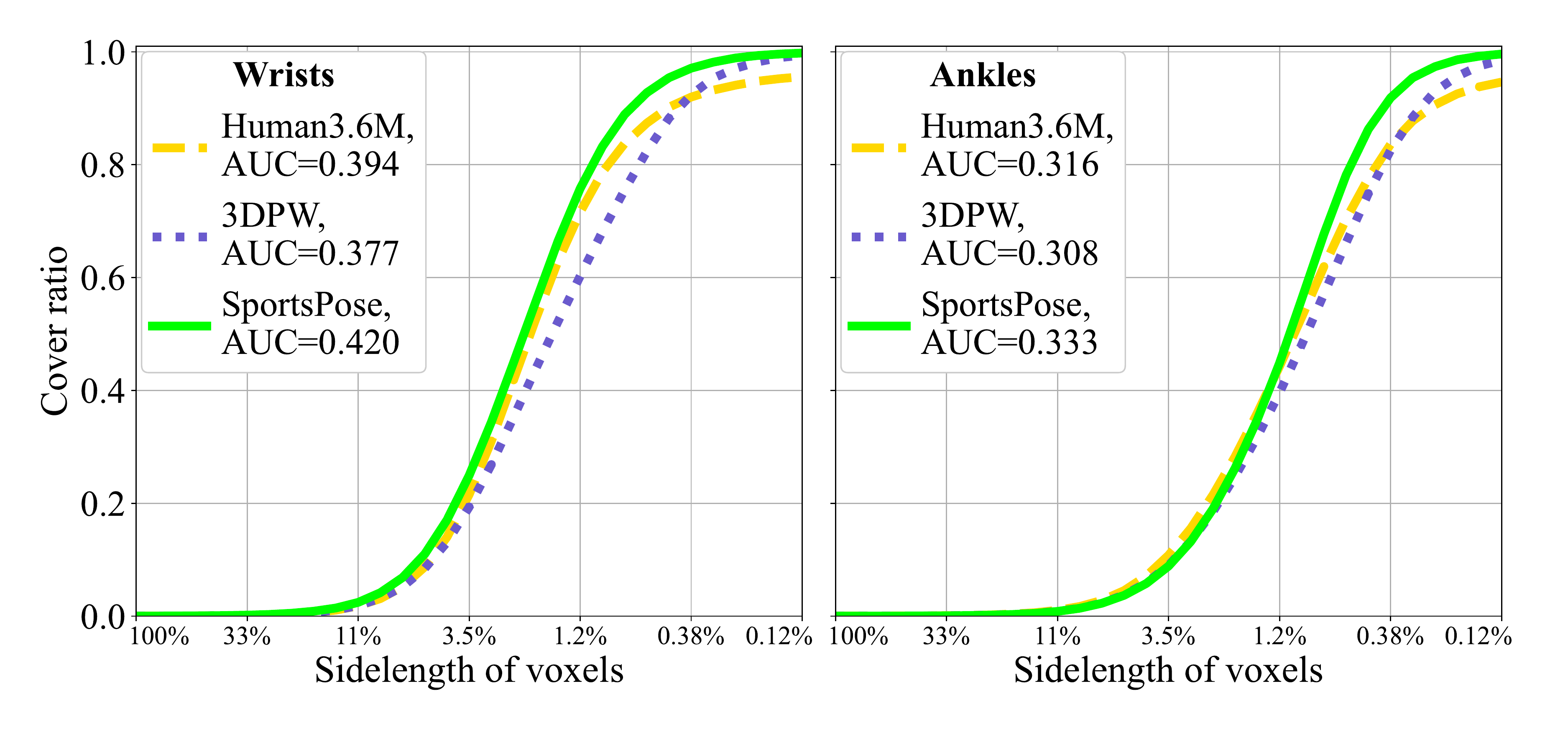}
  \caption{Fraction of unique voxels covered to the number of joints in the local coordinate system as a function of the size of the voxels for a subset of 50,000 poses from each dataset. The plot is a measure of how much movement is present in the different datasets.}
  \label{fig:LM_plot_50K}
\end{figure}

The local movement measure is sensitive to the number of poses used, and so to compare the datasets, we use a random subset of $50,000$ poses ($100,000$ joints including symmetry) for each of the three datasets considered. The resulting local movement metric is shown in \cref{fig:LM_plot_50K}. Selecting random subsets of poses can be done for this metric since we are not directly considering the movements, but only the poses that result from them, relaxing constraints on the order and origin of the individual poses. We see that SportsPose has the highest AUC for both wrists and ankles, indicating that our dataset includes a larger range of movements than both Human3.6M and 3DPW.

\section{Data quality assessment}
\label{sec:data_quality}
\subsection{Evaluation setup}
To verify the accuracy of our proposed markerless motion capture system, we have compared it to a commercial motion capture system from Qualisys \cite{qualisys}. The Qualisys system used is one of their most accurate systems consisting of eight Arqus A5 sensors and two Miqus Video cameras. To perform the evaluation, we connected both systems to a master synchronization unit, triggering both Qualisys and our markerless camera setup simultaneously at 90 Hz. This means that the two systems are frame synchronized and we can be certain that no measured discrepancies between our predicted poses and the Qualisys ground truth poses are caused by a time shift in the captures. 

 With our markerless system, we cannot freely choose the joints or points of interest to track. Here we are constrained to the joints detected with the used 2D pose detector which we use to triangulate the 3D joint positions. 2D pose detectors are trained on 2D datasets with joint labels annotated by humans without any biomechanical knowledge \cite{mpi2dpose, cocodataset}. This introduces some bias to the predictions, but assuming all annotations are correct, the annotators are asked to annotate the point corresponding to the joint center, which is a position inside the body. This is obviously not possible so the corresponding point on the surface of the body is instead annotated. 

\begin{figure}
    \centering
    \includegraphics[width=0.96\linewidth]{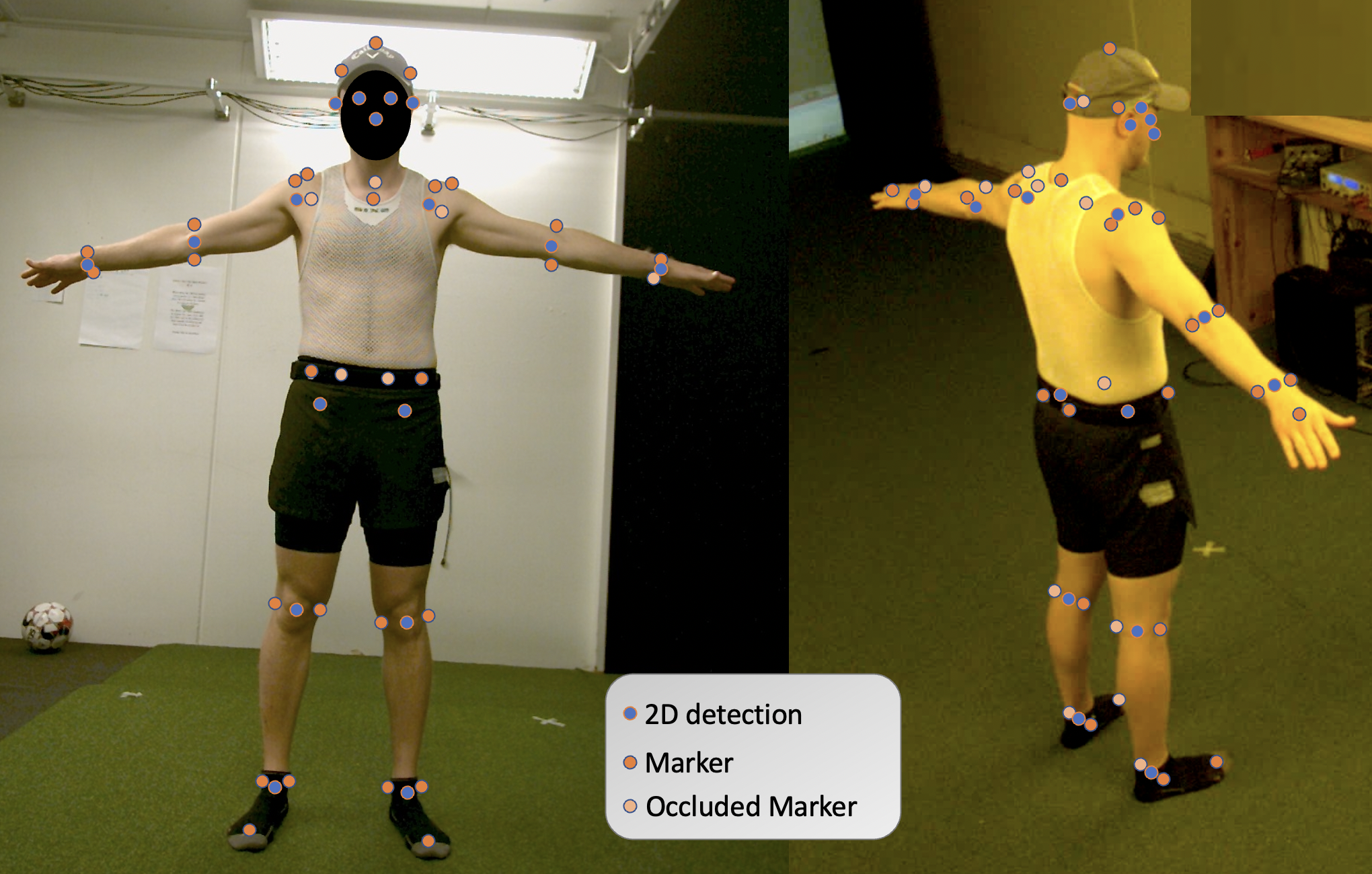}
    \caption{2D visualization of the detected joint centers from HRNet \cite{hrnet} in blue and the markers for the marker-based system \cite{qualisys} in orange with occluded markers shown in a lighter orange. It can be seen that the Qualisys markers are on the surface of the skin while the predicted HRNet is positioned in the joint center. From this illustration, it is clear that there is an offset between the two marker sets. }
    \label{fig:markerset}
\end{figure}

On the contrary, with the marker-based system, we can place markers on any joint or location on the body that we want to track freely. Ideally, to match the triangulated points from SportsPose the markers should be placed in the actual joint centers, which again is not possible because it is inside the body and we can only place markers on the surface of the body. We could place the markers in the same positions the markerless system detects but this depends on the viewpoint and triangulates to the actual joint center, where the marker-based system measures the actual marker location in 3D space. To overcome this we place the markers on anatomical landmarks on the body and use those locations to derive the actual joint center location \cite{dumas2007adjustments, wu2002isb, stagni2000effects}. The used anatomical landmarks are shown in \cref{fig:markerset} and are positioned directly on a rigid bone where possible. 

The captured sequences used for the quality assesment are not included in the $177$K poses, we present in \cref{tab:dataset_summary} but are $15,386$ additional poses, which also will be released as a seperate quality assesment subset of the data. The reason for this separation of our datasets is that we do not want new models to be trained on data where the subjects have visible markers attached. We do however assume that the attached markers wont benefit our quality assessment as the 2D pose detector is trained on the COCO dataset~\cite{cocodataset}, where there are no visible markers on the subjects.

For our comparison of the two systems, we used clothed subjects with tight clothing in order to minimize any movements between marker and joint, see \cref{fig:markerset}. We used clothed subjects to keep the setting as close to a real scenario as possible. With our markerset, illustrated in \cref{fig:markerset}, we ended up having most markers placed directly on the skin of the subject. The exceptions to this were markers placed on a belt around the hips, one set of markers on the shoes, and three markers around the head attached to a hat.%

\subsection{Aligning joint protocols}
\label{sec:align_joint}
\Cref{fig:markerset} shows the estimated joint locations from our markerless system and the corresponding Qualisys marker positions. From the figure, it is clear that there is a discrepancy between the two marker protocols, which we need to compensate for before doing the quality assessment of our motion capture system.

To compensate for the offset we have, for each subject in the evaluation capture, recorded a series of calibration recordings where the subject is either standing in a static pose with the arms out to the side as in \cref{fig:markerset} or performing a few slow and controlled movements. These sequences are used to compute a linear transformation from the markerset of the marker-based system to the markerset of our markerless system. \\

For each joint at time $t$ in the markerless system, $J^{(t)}$, we define a local joint coordinate system from three marker locations from the marker-based system, $M_1^{(t)}, M_2^{(t)}, M_3^{(t)}$. Two of them are the closest two markers to the joint and the last marker is chosen such that the joint moves little in relation to the plane spanned by the three locations. %

From the marker locations, we define the basis of the new local joint coordinate system as,

\begin{equation}
\begin{aligned}
    v_1^{(t)} &=  M_2^{(t)} - M_1^{(t)}\\
    v_2^{(t)} &=  M_3^{(t)} - M_1^{(t)}\\
    v_3^{(t)} &= v_1^{(t)} \times v_2^{(t)},
\end{aligned}
\end{equation}
\noindent
and set up the equation,
\begin{equation}
    A^{(t)}\,w + M_1^{(t)}= J^{(t)} \ .\\
\end{equation}
\noindent
Where, $w$ are the weights corresponding to the linear transformation, and $A^{(t)}$ contains the basis vectors of the local coordinate system,
\begin{equation}
    A^{(t)} = \begin{bmatrix} \frac{v_1^{(t)}}{\lVert v_1^{(t)}\rVert_2} &\frac{v_2^{(t)}}{\lVert v_2^{(t)}\rVert_2} &\frac{v_3^{(t)}}{\lVert v_3^{(t)}\rVert_2}\end{bmatrix}.
\end{equation}

Doing this for every timestep, $t$, and stacking all of the matrices into $A$, $J$, and $M_1$, we get one big system of equations, where we can compute the linear transformation by 

\begin{equation}
    w = (A^\top A)^{-1}A^\top (J - M_1).
\end{equation}

The estimated weights for the transformation, $w\in \mathbb{R}^3$, are unique for each joint for each subject and are used to transform the joints from the marker-based system to our markerless system.

\subsection{Quality assessment}
The quantitative quality assessment is done for two subjects who also are part of the main SportsPose dataset. The evaluation capture is done in the same physical location as the markerless indoor data and thus has identical lighting and background conditions. For each of the subjects, a series of calibration sequences were captured in order to learn the transformation between the markersets as described in \cref{sec:align_joint}. The calibration sequences are only used to estimate the transformations, and the evaluation is carried out on five repetitions of the five SportsPose activities of soccer, volleyball, jump, baseball pitch, and tennis. All of these sequences are converted to SportsPoses' markerset according to \cref{eq:gt_joints}.
\begin{equation}
    \Tilde{J}^{(t)} = A^{(t)} \, w + M_1^{(t)},
    \label{eq:gt_joints}
\end{equation}
\noindent
where $\Tilde{J}^{(t)}$ is the ground truth joint at time $t$. For the evaluation we adopted the evaluation protocol from Ingwersen et al.\ \cite{ingwersen2023evaluating}, i.e.\ computing the errors as, 
\begin{equation}
    \frac{1}{n} \sum_{t=1}^{n} \left|\left|\tau(J^{(t)}) - \Tilde{J}^{(t)} \right|\right|_2,
    \label{eq:eval}
\end{equation}
\noindent
where $\tau$ depends on the reported metric. This is calculated and averaged over the different joints. For the mean error in \cref{tab:quality_assessment}, ~$\tau$, is the identity transformation, for MPJPE it is a hip alignment and for PA-MPJPE, it is a full similarity transformation found by Procrustes analysis \cite{procrustes}. \\

The evaluation is done over all seventeen joints in the SportsPose dataset and the results can be found in \cref{tab:quality_assessment}. The mean error is 34.5~mm across all evaluation sequences. This shows that our ground truth is highly accurate for all movements in our dataset. Jumping has the highest error of the activities, still with a mean error below $4$~cm. From the table, we can also see that the mean error is lower than the hip aligned version and as expected the Procrustes aligned error is the lowest. A higher error after hip alignment suggests that we even after the alignment of the joint protocols described in \cref{sec:align_joint} still have an offset between the two protocols. However, since the Procrustes aligned error is lower this suggests that the majority of the remaining offset is in the location of the hips in the two protocols.

\begin{table}[]
\centering
\begin{tabular}{l*{3}{S[table-format=2.1]}}
\toprule
\text{Sequence} & \text{Mean error} & \text{MPJPE} & \text{PA-MPJPE} \\
\midrule
Baseball pitch & 36.5 & 42.6 & 30.5 \\ %
Jump & 38.4 & 48.2 & 35.9 \\ %
Tennis & 31.4 & 35.5 & 24.9 \\ %
Volleyball & 34.1 & 38.2 & 27.7 \\ %
Soccer & 32.0 & 30.0 & 26.5 \\ %
\midrule
Total & 34.5 & 38.9 & 29.1 \\ %
\bottomrule
\end{tabular}%
\caption{Quality assessment of the SportsPose dataset. It can be seen that the MPJPE is higher than the mean error which suggests that there is an offset between the SportsPose and marker-based systems hip location, while the other joints are fairly similar. Overall we have a comparable error to the 3DPW dataset \cite{3dpw}. All the errors are in mm.}
\label{tab:quality_assessment}
\end{table}

\section{Conclusion}
With SportsPose we provide the second largest publicly available markerless 3D human pose dataset in terms of poses and the dataset with the largest amount of subjects. The focus with the SportsPose dataset has been, as the name suggests, to capture a dataset with sports poses which naturally also are high-speed dynamic movements contrary to the poses seen in most other available datasets.

In addition to the number of subjects in the dataset, we also distinguish us from other markerless datasets through the thorough evaluation of the precision of our data with a commercial marker-based system. SportsPose is the only publicly available dataset where ground truth evaluation has been performed on data from the same domain as the data in the dataset. An additional advantage of working without markers is the lower set up time which has allowed us to make a diverse dataset with a large number of poses and subjects in multiple environments.

Our evaluation showed an average error of 34.5~mm which is comparable to the error reported by the commonly used 3DPW dataset \cite{3dpw}. We did however see that the Procrustes aligned error was lower than the hip aligned error which suggests that even after the alignment described in \cref{sec:align_joint} there is an offset in the position of the hip. 

We hope the SportsPose dataset is able to advance research and aid in development of methods for 3D human pose estimation methods that generalize better to faster and more extreme human movements.

{\small
\bibliographystyle{ieee_fullname}
\bibliography{egbib}

\begin{thebibliography}{10}\itemsep=-1pt

\bibitem{mpi2dpose}
Mykhaylo Andriluka, Leonid Pishchulin, Peter Gehler, and Bernt Schiele.
\newblock 2d human pose estimation: New benchmark and state of the art
  analysis.
\newblock In {\em IEEE Conference on Computer Vision and Pattern Recognition
  (CVPR)}, June 2014.

\bibitem{smplify}
Federica Bogo, Angjoo Kanazawa, Christoph Lassner, Peter Gehler, Javier Romero,
  and Michael~J Black.
\newblock {Keep It SMPL: Automatic Estimation of 3D Human Pose and Shape from a
  Single Image BT - Computer Vision – ECCV 2016}.
\newblock pages 561--578, Cham, 2016. Springer International Publishing.

\bibitem{CapturyMarkerless}
Markerless computer vision tracking.
\newblock \url{https://captury.com/}, 2022.

\bibitem{chen2020anatomy}
Tianlang Chen, Chen Fang, Xiaohui Shen, Yiheng Zhu, Zhili Chen, and Jiebo Luo.
\newblock Anatomy-aware 3d human pose estimation in videos.
\newblock {\em arXiv preprint arXiv:2002.10322}, 2020.

\bibitem{dumas2007adjustments}
Rapha{\"e}l Dumas, Laurence Cheze, and J-P Verriest.
\newblock Adjustments to mcconville et al. and young et al. body segment
  inertial parameters.
\newblock {\em Journal of biomechanics}, 40(3):543--553, 2007.

\bibitem{procrustes}
J.~C. Gower.
\newblock {Generalized procrustes analysis}.
\newblock {\em Psychometrika}, 40(1):33--51, 1975.

\bibitem{dynaboa}
Shanyan Guan, Jingwei Xu, Michelle~Z. He, Yunbo Wang, Bingbing Ni, and Xiaokang
  Yang.
\newblock Out-of-domain human mesh reconstruction via dynamic bilevel online
  adaptation.
\newblock {\em IEEE Transactions on Pattern Analysis and Machine Intelligence},
  pages 1--16, 2022.

\bibitem{hwang2017a}
Jihye Hwang, Sungheon Park, and Nojun Kwak.
\newblock Athlete pose estimation by a global-local network.
\newblock {\em Ieee Computer Society Conference on Computer Vision and Pattern
  Recognition Workshops}, 2017-:114--121, 2017.

\bibitem{ingwersen2023evaluating}
Christian~Keilstrup Ingwersen, Janus~N{\o}rtoft Jensen, Morten~Rieger
  Hannemose, and Anders~B. Dahl.
\newblock Evaluating current state of monocular 3d pose models for golf.
\newblock In {\em Proceedings of the Northern Lights Deep Learning Workshop},
  volume~4, 2023.

\bibitem{h36m}
Catalin Ionescu, Dragos Papava, Vlad Olaru, and Cristian Sminchisescu.
\newblock Human3.6m: Large scale datasets and predictive methods for 3d human
  sensing in natural environments.
\newblock {\em IEEE Transactions on Pattern Analysis and Machine Intelligence},
  36(7):1325--1339, 2014.

\bibitem{panoptic}
Hanbyul Joo, Hao Liu, Lei Tan, Lin Gui, Bart Nabbe, Iain Matthews, Takeo
  Kanade, Shohei Nobuhara, and Yaser Sheikh.
\newblock Panoptic studio: A massively multiview system for social motion
  capture.
\newblock In {\em 2015 {IEEE} International Conference on Computer Vision
  ({ICCV})}. {IEEE}, dec 2015.

\bibitem{hmr}
Angjoo Kanazawa, Michael~J. Black, David~W. Jacobs, and Jitendra Malik.
\newblock End-to-end recovery of human shape and pose.
\newblock In {\em 2018 IEEE/CVF Conference on Computer Vision and Pattern
  Recognition}, pages 7122--7131, 2018.

\bibitem{kocabas2021pare}
Muhammed Kocabas, Chun-Hao~P. Huang, Otmar Hilliges, and Michael~J. Black.
\newblock Pare: Part attention regressor for 3d human body estimation.
\newblock In {\em 2021 IEEE/CVF International Conference on Computer Vision
  (ICCV)}, pages 11107--11117, 2021.

\bibitem{spin}
Nikos Kolotouros, Georgios Pavlakos, Michael Black, and Kostas Daniilidis.
\newblock Learning to reconstruct 3d human pose and shape via model-fitting in
  the loop.
\newblock In {\em 2019 IEEE/CVF International Conference on Computer Vision
  (ICCV)}, pages 2252--2261, 2019.

\bibitem{metro}
Kevin Lin, Lijuan Wang, and Zicheng Liu.
\newblock End-to-end human pose and mesh reconstruction with transformers.
\newblock In {\em 2021 IEEE/CVF Conference on Computer Vision and Pattern
  Recognition (CVPR)}, pages 1954--1963, 2021.

\bibitem{meshgraphformer}
Kevin Lin, Lijuan Wang, and Zicheng Liu.
\newblock Mesh graphormer.
\newblock In {\em 2021 IEEE/CVF International Conference on Computer Vision
  (ICCV)}, pages 12919--12928, 2021.

\bibitem{cocodataset}
Tsung{-}Yi Lin, Michael Maire, Serge~J. Belongie, Lubomir~D. Bourdev, Ross~B.
  Girshick, James Hays, Pietro Perona, Deva Ramanan, Piotr Doll{'{a} }r, and
  C.~Lawrence Zitnick.
\newblock Microsoft {COCO:} common objects in context.
\newblock {\em CoRR}, abs/1405.0312, 2014.

\bibitem{Loper:SIGASIA:2014}
Matthew~M. Loper, Naureen Mahmood, and Michael~J. Black.
\newblock {MoSh}: Motion and shape capture from sparse markers.
\newblock {\em ACM Transactions on Graphics, (Proc. SIGGRAPH Asia)},
  33(6):220:1--220:13, Nov. 2014.

\bibitem{3dhp}
Dushyant Mehta, Helge Rhodin, Dan Casas, Pascal Fua, Oleksandr Sotnychenko,
  Weipeng Xu, and Christian Theobalt.
\newblock Monocular 3d human pose estimation in the wild using improved cnn
  supervision.
\newblock In {\em 2017 International Conference on 3D Vision (3DV)}, pages
  506--516, 2017.

\bibitem{mirek2007assessment}
Elzbieta Mirek, Monika Rudzi{\'n}ska, and Andrzej Szczudlik.
\newblock The assessment of gait disorders in patients with parkinson's disease
  using the three-dimensional motion analysis system vicon.
\newblock {\em Neurologia i neurochirurgia polska}, 41(2):128--133, 2007.

\bibitem{aspset}
Aiden Nibali, Joshua Millward, Zhen He, and Stuart Morgan.
\newblock {ASPset}: An outdoor sports pose video dataset with 3d keypoint
  annotations.
\newblock {\em Image and Vision Computing}, 111:104196, jul 2021.

\bibitem{nishimura2020a}
Makoto Nishimura, Makiko Itoi, Masaki Saito, Kensuke Tsurumaki, Miki Kurushima,
  and Kiyoko Tokunaga.
\newblock Nursing students' motion posture evaluation using human pose
  estimation.
\newblock {\em International Journal of Learning}, 6(1):43--46, 2020.

\bibitem{parks1987a}
Thomas~W. Parks and Charles~S. Burrus.
\newblock {\em Digital filter design}.
\newblock Wiley, 1987.

\bibitem{pavllo:videopose3d:2019}
Dario Pavllo, Christoph Feichtenhofer, David Grangier, and Michael Auli.
\newblock 3d human pose estimation in video with temporal convolutions and
  semi-supervised training.
\newblock In {\em Conference on Computer Vision and Pattern Recognition
  (CVPR)}, 2019.

\bibitem{qualisys}
Motion capture by qualisys.
\newblock \url{https://www.qualisys.com/}, 2022.

\bibitem{rematas2018a}
Konstantinos Rematas, Ira Kemelmacher-Shlizerman, Brian Curless, and Steve
  Seitz.
\newblock Soccer on your tabletop.
\newblock {\em Proceedings of the Ieee Computer Society Conference on Computer
  Vision and Pattern Recognition}, pages 4738--4747, 2018.

\bibitem{aruco}
Francisco~J. Romero-Ramirez, Rafael Muñoz-Salinas, and Rafael Medina-Carnicer.
\newblock Speeded up detection of squared fiducial markers.
\newblock {\em Image and Vision Computing}, 76:38--47, 2018.

\bibitem{ronchi2017a}
Matteo~Ruggero Ronchi and Pietro Perona.
\newblock Benchmarking and error diagnosis in multi-instance pose estimation.
\newblock 2017.

\bibitem{sandbakk2012influence}
{\O}yvind Sandbakk, Gertjan Ettema, and Hans-Christer Holmberg.
\newblock The influence of incline and speed on work rate, gross efficiency and
  kinematics of roller ski skating.
\newblock {\em European journal of applied physiology}, 112:2829--2838, 2012.

\bibitem{scott2017a}
Jesse Scott, Robert Collins, Christopher Funk, and Yanxi Liu.
\newblock 4d model-based spatiotemporal alignment of scripted taiji quan
  sequences.
\newblock {\em Proceedings - 2017 Ieee International Conference on Computer
  Vision Workshops, Iccvw 2017}, 2018-:795--804, 2017.

\bibitem{shan2022p}
Wenkang Shan, Zhenhua Liu, Xinfeng Zhang, Shanshe Wang, Siwei Ma, and Wen Gao.
\newblock P-stmo: Pre-trained spatial temporal many-to-one model for 3d human
  pose estimation.
\newblock In {\em Computer Vision--ECCV 2022: 17th European Conference, Tel
  Aviv, Israel, October 23--27, 2022, Proceedings, Part V}, pages 461--478.
  Springer, 2022.

\bibitem{HumanEva}
Leonid Sigal, Alexandru~O Balan, and Michael~J Black.
\newblock {HumanEva: Synchronized Video and Motion Capture Dataset and Baseline
  Algorithm for Evaluation of Articulated Human Motion}.
\newblock {\em International Journal of Computer Vision}, 87(1):4, 2009.

\bibitem{simiMarkerless}
Markerless motion capture for every application: Simi.
\newblock \url{https://simishape.com/}, 2022.

\bibitem{stagni2000effects}
Rita Stagni, Alberto Leardini, Aurelio Cappozzo, Maria~Grazia Benedetti, and
  Angelo Cappello.
\newblock Effects of hip joint centre mislocation on gait analysis results.
\newblock {\em Journal of biomechanics}, 33(11):1479--1487, 2000.

\bibitem{hrnet}
Ke Sun, Bin Xiao, Dong Liu, and Jingdong Wang.
\newblock Deep high-resolution representation learning for human pose
  estimation.
\newblock In {\em Proceedings of the IEEE/CVF conference on computer vision and
  pattern recognition}, pages 5693--5703, 2019.

\bibitem{romp}
Yu Sun, Qian Bao, Wu Liu, Yili Fu, Michael~J. Black, and Tao Mei.
\newblock Monocular, one-stage, regression of multiple 3d people.
\newblock In {\em 2021 IEEE/CVF International Conference on Computer Vision
  (ICCV)}, pages 11159--11168, 2021.

\bibitem{TheiaMarkerless}
Markerless motion capture redefined.
\newblock \url{https://www.theiamarkerless.ca/}, 2022.

\bibitem{totalcapture}
Matthew Trumble, Andrew Gilbert, Charles Malleson, Adrian Hilton, and John
  Collomosse.
\newblock Total capture: 3d human pose estimation fusing video and inertial
  sensors.
\newblock In Gabriel~Brostow Tae-Kyun~Kim, Stefanos~Zafeiriou and Krystian
  Mikolajczyk, editors, {\em Proceedings of the British Machine Vision
  Conference (BMVC)}, pages 14.1--14.13. BMVA Press, September 2017.

\bibitem{3dpw}
Timo von Marcard, Roberto Henschel, Michael~J. Black, Bodo Rosenhahn, and
  Gerard Pons-Moll.
\newblock Recovering accurate 3d human pose in the wild using imus and a moving
  camera.
\newblock In {\em Proceedings of the European Conference on Computer Vision
  (ECCV)}, September 2018.

\bibitem{williams2006a}
Arthur~Bernard Williams.
\newblock {\em Electronic filter design handbook, Electronic filter design
  handbook, 4th ed}.
\newblock McGraw-Hill, 2006.

\bibitem{winter2009a}
David~A. Winter.
\newblock {\em Biomechanics and Motor Control of Human Movement: Fourth
  Edition}.
\newblock John Wiley and Sons, 2009.

\bibitem{winter1974a}
David~A Winter, H~Grant Sidwall, and Douglas~A Hobson.
\newblock Measurement and reduction of noise in kinematics of locomotion.
\newblock {\em Journal of biomechanics}, 7(2):157--159, 1974.

\bibitem{wu2002isb}
Ge Wu, Sorin Siegler, Paul Allard, Chris Kirtley, Alberto Leardini, Dieter
  Rosenbaum, Mike Whittle, Darryl D~D’Lima, Luca Cristofolini, Hartmut Witte,
  et~al.
\newblock Isb recommendation on definitions of joint coordinate system of
  various joints for the reporting of human joint motion—part i: ankle, hip,
  and spine.
\newblock {\em Journal of biomechanics}, 35(4):543--548, 2002.

\bibitem{xu2020ghum}
Hongyi Xu, Eduard~Gabriel Bazavan, Andrei Zanfir, William~T Freeman, Rahul
  Sukthankar, and Cristian Sminchisescu.
\newblock Ghum \& ghuml: Generative 3d human shape and articulated pose models.
\newblock In {\em Proceedings of the IEEE/CVF Conference on Computer Vision and
  Pattern Recognition}, pages 6184--6193, 2020.

\bibitem{zecha2018a}
Dan Zecha, Moritz Einfalt, Christian Eggert, and Rainer Lienhart.
\newblock Kinematic pose rectification for performance analysis and retrieval
  in sports.
\newblock {\em Ieee Computer Society Conference on Computer Vision and Pattern
  Recognition Workshops}, 2018-:1872--1880, 2018.

\bibitem{Zhang_2022_CVPR}
Jinlu Zhang, Zhigang Tu, Jianyu Yang, Yujin Chen, and Junsong Yuan.
\newblock Mixste: Seq2seq mixed spatio-temporal encoder for 3d human pose
  estimation in video.
\newblock In {\em Proceedings of the IEEE/CVF Conference on Computer Vision and
  Pattern Recognition (CVPR)}, pages 13232--13242, June 2022.

\bibitem{zhang2000a}
Zhengyou Zhang.
\newblock A flexible new technique for camera calibration.
\newblock {\em Ieee Transactions on Pattern Analysis and Machine Intelligence},
  22(11):1330--1334, 2000.

\bibitem{zhang2012a}
Zhengyou Zhang.
\newblock Microsoft kinect sensor and its effect.
\newblock {\em Ieee Multimedia}, 19(2):4--10, 2012.

\end{thebibliography}
}

\end{document}